\documentclass{article}
\usepackage{ijcai20}
\usepackage{times}
\usepackage{soul}
\usepackage{url}
\usepackage[hidelinks]{hyperref}
\usepackage[utf8]{inputenc}
\usepackage[small]{caption}
\usepackage{graphicx}
\usepackage{amsmath}
\usepackage{amsthm}
\usepackage{amssymb}
\usepackage{bm}
\usepackage{booktabs}
\usepackage{algorithm}
\usepackage{algorithmic}
\usepackage{multirow}
\usepackage{latexsym}

\newcommand{\R}{\mathbb{R}}
\newcommand{\Gcal}{\mathcal{G}}
\newcommand{\Vcal}{\mathcal{V}}
\newcommand{\Ecal}{\mathcal{E}}
\newcommand{\Ocal}{\mathcal{O}}
\newcommand{\Ncal}{\mathcal{N}}

\DeclareMathOperator*{\softmax}{softmax}

\title{Mucko: Multi-Layer Cross-Modal Knowledge Reasoning for Fact-based Visual Question Answering}

\author{
Zihao Zhu$^{1,2}$\footnote{Equal contribution.}\and
Jing Yu$^{1,2*}$\and
Yujing Wang$^{3}$\and
Yajing Sun$^{1,2}$\and
Yue Hu$^{1,2}$\And
Qi Wu $^{4}$\\
\affiliations
$^1$Institute of Information Engineering, Chinese Academy of Sciences, Beijing, China\\
$^2$School of Cyber Security, University of Chinese Academy of Sciences, Beijing, China\\
$^3$Microsoft Research Asia, Beijing, China\\
$^4$University of Adelaide, Australia\\
\emails
\{zhuzihao, yujing02, sunyajing, huyue\}@iie.ac.cn,
yujwang@microsoft.com,
qi.wu01@adelaide.edu.au
}

\begin{document}
\maketitle
\begin{abstract}
Fact-based Visual Question Answering (FVQA) requires external knowledge beyond visible content to answer questions about an image, which is challenging but indispensable to achieve general VQA. One limitation of existing FVQA solutions is that they jointly embed all kinds of information without fine-grained selection, which introduces unexpected noises for reasoning the final answer. How to capture the question-oriented and information-complementary evidence remains a key challenge to solve the problem.
In this paper, we depict an image by a multi-modal heterogeneous graph, which contains multiple layers of information corresponding to the visual, semantic and factual features.
On top of the multi-layer graph representations, we propose a modality-aware heterogeneous graph convolutional network to capture evidence from different layers that is most relevant to the given question. Specifically, the intra-modal graph convolution selects evidence from each modality and cross-modal graph convolution aggregates relevant information across different modalities. 
By stacking this process multiple times, our model performs iterative reasoning and predicts the optimal answer by analyzing all question-oriented evidence. We achieve a new state-of-the-art performance on the FVQA task and demonstrate the effectiveness and interpretability of our model with extensive experiments.
\end{abstract}

\section{Introduction}
\begin{figure}[t]
	\centering
	\setlength{\abovecaptionskip}{5pt}
	\setlength{\belowcaptionskip}{-5mm}
	\includegraphics[width=\columnwidth]{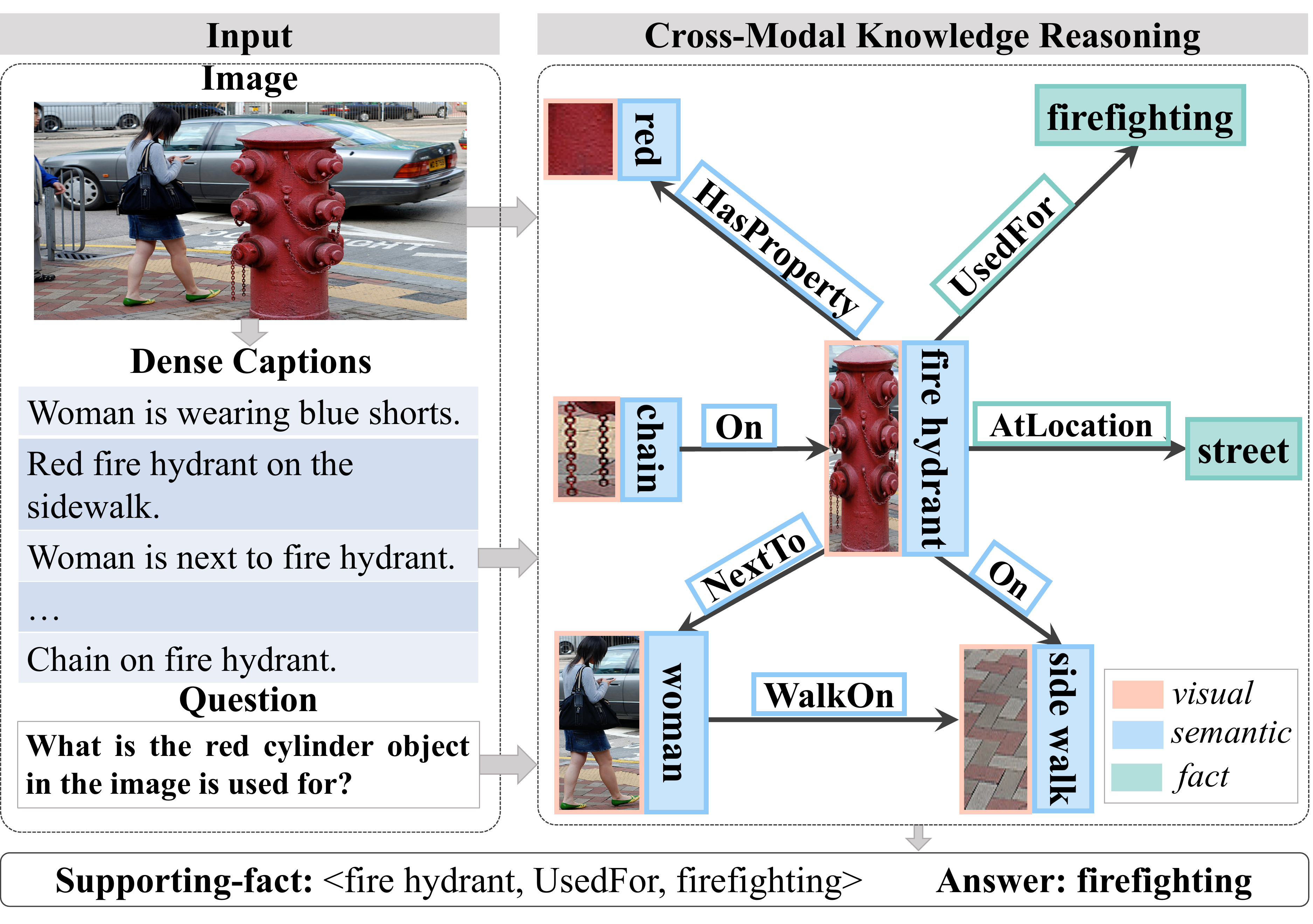}
	\caption{\small{An illustration of our motivation. We represent an image by multi-layer graphs and cross-modal knowledge reasoning is conducted on the graphs to infer the optimal answer.}}
	\label{fig:intro}
\end{figure}

Visual question answering (VQA) \cite{antol2015vqa} is an attractive research direction aiming to jointly analyze multimodal content from images and natural language. Equipped with the capacities of grounding, reasoning and translating, a VQA agent is expected to answer a question in natural language based on an image. Recent works~\cite{cadene2019murel,li2019relation,ben2019block} have achieved great success in the VQA problems that are answerable by solely referring to the visible content of the image. However, such kinds of models are incapable of answering questions which require external knowledge beyond what is in the image. 
Considering the question in Figure \ref{fig:intro}, the agent not only needs to visually localize `the red cylinder', but also to semantically recognize it as `fire hydrant' and connects the knowledge that `fire hydrant is used for firefighting'. 
Therefore, how to collect the question-oriented and information-complementary evidence from visual, semantic and knowledge perspectives is essential to achieve general VQA.

To advocate research in this direction, ~\cite{wang2018fvqa} introduces the `Fact-based' VQA (FVQA) task for answering questions by joint analysis of the image and the knowledge base of facts. The typical solutions for FVQA build a fact graph with fact triplets filtered by the visual concepts in the image 
and select one entity in the graph as the answer. 
Existing works \cite{wang2017explicit,wang2018fvqa} parse the question as keywords and retrieve the supporting-entity only by keyword matching. This kind of approaches is vulnerable when the question does not exactly mention the visual concepts (\textit{e.g.} synonyms and homographs) or the mentioned information is not captured in the fact graph (\textit{e.g.} the visual attribute `red' in Figure \ref{fig:intro} may be falsely omitted). To resolve these problems, \cite{narasimhan2018out} introduces visual information into the fact graph and infers the answer by implicit graph reasoning under the guidance of the question. However, they provide the whole visual information equally to each graph node by concatenation of the image, question and entity embeddings. Actually, only part of the visual content are relevant to the question and a certain entity.
Moreover, the fact graph here is still homogeneous since each node is represented by a fixed form of image-question-entity embedding, which limits the model's flexibility of adaptively capturing evidence from different modalities. 

In this work, we depict an image as a multi-modal heterogeneous graph, which contains multiple layers of information corresponding to different modalities. 
The proposed model is focused on \textit{\textbf{Mu}lti-Layer \textbf{C}ross-Modal \textbf{K}nowledge Reas\textbf{o}ning} 
and we name it as \textbf{Mucko} for short. Specifically, we encode an image by three layers of graphs, where the object appearance and their relationships are kept in the \textit{visual layer}, the high-level abstraction for bridging the gaps between visual and factual information is provided in the \textit{semantic layer}, and the corresponding knowledge of facts are supported in the \textit{fact layer}. We propose a modality-aware heterogeneous graph convolutional network to adaptively collect complementary evidence in the multi-layer graphs. It can be performed by two procedures. First, the Intra-Modal Knowledge Selection procedure collects question-oriented information from each graph layer under the guidance of question; Then, the Cross-Modal Knowledge Reasoning procedure captures complementary evidence across different layers. 

The main contributions of this paper are summarized as follows:
(1) We comprehensively depict an image by a heterogeneous graph containing multiple layers of information based on visual, semantic and knowledge modalities. We consider these three modalities jointly and achieve significant improvement over state-of-the-art solutions. 
(2) We propose a modality-aware heterogeneous graph convolutional network to capture question-oriented evidence from different modalities. 
Especially, we leverage an attention operation in each convolution layer to select the most relevant evidence for the given question, and the convolution operation is responsible for adaptive feature aggregation.
(3) We demonstrate good interpretability of our approach and provide case study in deep insights. Our model automatically tells which modality (visual, semantic or factual) and entity have more contributions to answer the question through visualization of attention weights and gate values.

\section{Related Work}

\paragraph{Visual Question Answering.}

The typical solutions for VQA are based on the CNN-RNN architecture \cite{malinowski2015ask} and leverage global visual features to represent image, which may introduce noisy information.
Various attention mechanisms \cite{yang2016stacked,lu2016hierarchical,anderson2018bottom} have been exploited to highlight visual objects that are relevant to the question. However, they treat objects independently and ignore their informative relationships. 
\cite{battaglia2018relational} demonstrates that human’s ability of combinatorial generalization highly depends on the mechanisms for reasoning over relationships. 
Consistent with such proposal, there is an emerging trend to represent the image by graph structure to depict objects and relationships in VQA and other vision-language tasks \cite{hu2019language,wang2019neighbourhood,li2019relation}.
As an extension, \cite{jiang2019dualvd} exploits natural language to enrich the graph-based visual representations.
However, it solely captures the semantics in natural language by LSTM, which lacking of fine-grained correlations with the visual information.
To go one step further, we depict an image by multiple layers of graphs from visual, semantic and factual perspectives to collect fine-grained evidence from different modalities.

\paragraph{Fact-based Visual Question Answering.}
Human can easily combine visual observation with external knowledge for answering questions, which remains challenging for algorithms.
\cite{wang2018fvqa} introduces a fact-based VQA task, which provides a knowledge base of facts and associates each question with a supporting-fact.
Recent works based on FVQA generally select one entity from fact graph as the answer and falls into two categories: query-mapping based methods and learning based methods.
\cite{wang2017explicit} reduces the question to one of the available query templates and this limits the types of questions that can be asked. 
\cite{wang2018fvqa} automatically classifies and maps the question to a query which does not suffer the above constraint.
Among both methods, however, visual information are used to extract facts but not introduced during the reasoning process.
~\cite{narasimhan2018out} applies GCN on the fact graph where each node is represented by the fixed form of image-question-entity embedding. However, the visual information is wholly provided which may introduce redundant information for prediction.
In this paper, we decipt an image by multi-layer graphs and perform cross-modal heterogeneous graph reasoning on them to capture complementary evidence from different layers that most relevant to the question.

\paragraph{Heterogeneous Graph Neural Networks.}
Graph neural networks  are gaining fast momentum in the last few years~\cite{wu2019comprehensive}.
Compared with homogeneous graphs, heterogeneous graphs are more common in the real world.
\cite{schlichtkrull2018modeling} generalizes graph convolutional network (GCN) to handle different relationships between entities in a knowledge base, where each edge with distinct relationships is encoded independently.
\cite{wang2019heterogeneous,hu2019heterogeneous} propose heterogeneous graph attention networks with dual-level attention mechanism.
All of these methods model different types of nodes and edges on a unified graph.
In contrast, the heterogeneous graph in this work contains multiple layers of subgraphs and each layer consists of nodes and edges coming from different modalities. For this specific constrain, we propose the intra-modal and cross-modal graph convolutions for reasoning over such multi-modal heterogeneous graphs.

\section{Methodology}
\begin{figure*}[t]
	\setlength{\abovecaptionskip}{5pt}
	\setlength{\belowcaptionskip}{-14pt}
	\includegraphics[width =\textwidth]{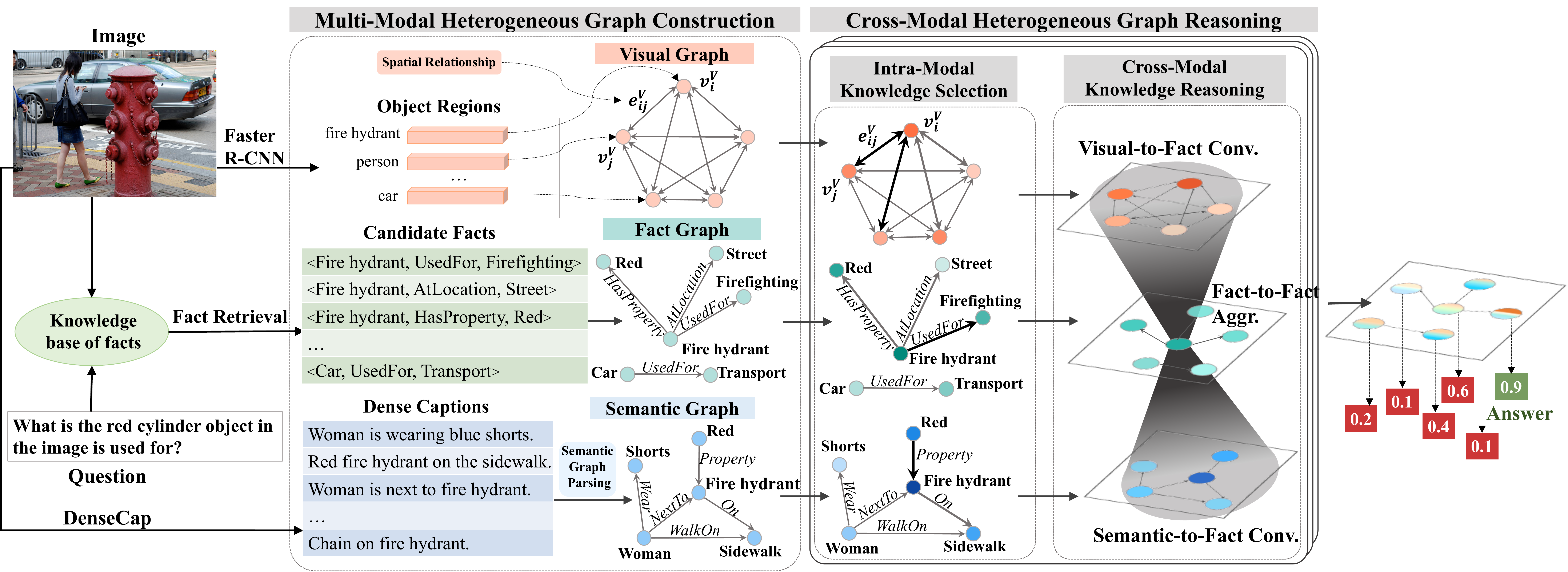}
	\caption{\small{An overview of our model. The model contains two modules: Multi-modal Heterogeneous Graph Construction aims to depict an image by multiple layers of graphs and Cross-modal Hetegeneous Graph Reasoning supports intra-modal and cross-modal evidence selection.
	}}
	\label{fig:model}
\end{figure*} 

Given an image $I$ and a question $Q$, the task aims to predict an answer $A$ while leveraging external knowledge base, which consists of facts in the form of triplet, \textit{i.e.} $<e_1, r, e_2>$, where $e_1$ is a visual concept in the image, $e_2$ is an attribute or phrase and $r$ represents the relationship between $e_1$ and $e_2$. 
The key is to choose a correct entity, \textit{i.e.} either $e_1$ or $e_2$, from the supporting fact as the predicted answer. We first introduce a novel scheme of depicting an image by three layers of graphs, including the visual graph, semantic graph and fact graph respectively, imitating the understanding of various properties of an object and the relationships. Then we perform cross-modal heterogeneous graph reasoning that consists of two parts: \textit{Intra-Modal Knowledge Selection} aims to choose question-oriented knowledge from each layer of graphs by intra-modal graph convolutions, and \textit{Cross-Modal Knowledge Reasoning} adaptively selects complementary evidence across three layers of graphs by cross-modal graph convolutions. 
By stacking the above two processes  multiple times, our model performs iterative reasoning across all the modalities and results in the optimal answer by jointly analyzing all the entities. Figure \ref{fig:model} gives detailed illustration of our model.

\subsection{Multi-Modal Graph Construction}\label{sec:graphConstruction}

\paragraph{Visual Graph Construction.}
Since most of the questions in FVQA grounded in the visual objects and their relationships, we construct a fully-connected visual graph to represent such evidence at appearance level. Given an image $I$, we use Faster-RCNN \cite{ren2015faster} to identify a set of objects $\Ocal = \{o_i\}_{i=1}^K$ ($K$ = 36), where each object $o_i$ is associated with a visual feature vector $\bm{v}_i \in \R^{d_v}$ ($d_v$ = 2048), a spatial feature vector $\bm{b}_i\in \R^{d_b}$ ($d_b$ = 4) and a corresponding label. 
Specifically, $\bm{b}_i = [x_i, y_i, w_i, h_i]$, where $(x_i, y_i)$, $h_i$ and $w_i$ respectively denote the coordinate of the top-left corner, the height and width of the bounding box.
We construct a visual graph $\Gcal^V=(\Vcal^V,\Ecal^V)$ over $\Ocal$, where $\Vcal^V =\{v_i^V\}_{i=1}^K$ is the node set and each node $v^V_{i}$ corresponds to a detected object $o_i$. 
The feature of node $v^{V}_{i}$ is represented by $\bm{v}^V_i$. Each edge $e^V_{ij} \in \Ecal^V$ denotes the relative spatial relationships between two objects. We encode the edge feature by a 5-dimensional vector, \textit{i.e.} $\bm{r}^V_{ij} = [\frac{x_j-x_i}{w_i},\frac{y_j-y_i}{h_i},\frac{w_j}{w_i},\frac{h_j}{h_i},\frac{w_jh_j}{w_ih_i}]$.

\paragraph{Semantic Graph Construction.}
In addition to visual information, high-level abstraction of the objects and relationships by natural language provides essential semantic information. Such abstraction is indispensable to associate the visual objects in the image with the concepts mentioned in both questions and facts. In our work, we leverage dense captions \cite{johnson2016densecap} to extract a set of local-level semantics in an image, ranging from the properties of a single object (color, shape, emotion, $etc.$) to the relationships between objects (action, spatial positions, comparison, $etc.$). We decipt an image by $D$ dense captions, denoted as $Z=\{z_i\}_{i=1}^D$, where $z_i$ is a natural language description about a local region in the image. Instead of using monolithic embeddings to represent the captions, we exploit to model them by a graph-based semantic representation, denoted as $\Gcal^S=(\Vcal^S,\Ecal^S)$, which is constructed by a semantic graph parsing model \cite{anderson2016spice}. 
The node $v^S_i \in \Vcal^S$ represents the name or attribute of an object extracted from the captions while the edge $e_{ij}^S\in \Ecal^S$ represents the relationship between $v^S_i$ and $v^S_j$. We use the averaged GloVe embeddings \cite{pennington2014glove} to represent $v^S_i$ and $e_{ij}^S$, denoted as $\bm{v}_{i}^S$ and $\bm{r}_{ij}^S$, respectively. The graph representation retains the relational information among concepts and unifies the representations in graph domain, which is better for explicit reasoning across modalities. 

\paragraph{Fact Graph Construction.}

To find the optimal supporting-fact, we first retrieve relevant candidate facts from knowledge base of facts following a scored based approach proposed in \cite{narasimhan2018out}. We compute the cosine similarity of the embeddings of every word in the fact with the words in the question and the words of visual concepts detected in the image. Then we average these values to assign a similarity score to the fact. The facts are sorted based on the similarity and the 100 highest scoring facts are retained, denoted as $f_{100}$. A relation type classifier is trained additionally to further filter the retrieved facts.
Specifically, we feed the last hidden state of LSTM to an MLP layer to predict the relation type $\hat{r}_i$ of a question. We retain the facts among $f_{100}$ only if their relationships agree with $\hat{r}_i$, \textit{i.e.} $f_{rel}=f\in f_{100}:r(f) \in\{\hat{r}_i\}$ ($\{\hat{r}_i\}$ contains top-3 predicted relationships in experiments). 
Then a fact graph $\Gcal^F=(\Vcal^F,\Ecal^F)$ is built upon $f_{rel}$ as the candidate facts can be naturally organized as graphical structure.
Each node $v_i^F \in \Vcal^F$ denotes an entity in $f_{rel}$ and is represented by GloVe embedding of the entity, denoted as $\bm{v}_i^F$.
Each edge $e_{ij}^F \in \Ecal^F$ denotes the relationship between $v_i^F$ and $v_j^F$ and is represented by GloVe embedding $\bm{r}_{ij}$. The topological structure among facts can be effectively exploited by jointly considering all the entities in the fact graph.

\subsection{Intra-Modal Knowledge Selection}
\label{subsec:Intra}
Since each layer of graphs contains modality-specific knowledge relevant to the question, we first select valuable evidence independently from the visual graph, semantic graph and fact graph by \textbf{Visual-to-Visual Convolution}, \textbf{Semantic-to-Semantic Convolution} and \textbf{Fact-to-Fact Convolution} respectively. These three convolutions share the common operations but differ in their node and edge representations corresponding to the graph layers. Thus we omit the superscript of node representation $\bm{v}$ and edge representation $\bm{r}$ in the rest of this section. 
We first perform attention operations to highlight the nodes and edges that are most relevant to the question $q$ and consequently update node representations via intra-modal graph convolution. 
This process mainly consists of the following three steps:

\paragraph{Question-guided Node Attention.}
We first evaluate the relevance of each node corresponding to the question by attention mechanism. The attention weight for $v_i$ is computed as:
\begin{equation}\label{eq:node attention}
\alpha_i=\softmax(\bm{w}^T_a\tanh({\textbf{W}_{1} }\bm{v}_i + \textbf{W}_{2}\bm{q}))
\end{equation}
where $\textbf{W}_{1}$,$\textbf{W}_{2}$ and $\bm{w}_a$ (as well as $\textbf{W}_{3}$,..., $\textbf{W}_{11}$, $\bm{w}_b$, $\bm{w}_c$ mentioned below) are learned parameters. $\bm{q}$ is question embedding encoded by LSTM.

\paragraph{Question-guided Edge Attention.} Under the guidance of question, we then evaluate the importance of edge $e_{ji}$ constrained by the neighbor node $v_j$ regarding to $v_i$ as:
\begin{equation}\label{edge attention}
\beta_{ji}=\softmax(\bm{w}^T_b\tanh({\textbf{W}_{3} }\bm{v'}_{j} + \textbf{W}_{4}\bm{q'}))
\end{equation}
where $\bm{v'}_{j}=\textbf{W}_5[\bm{v}_j,\bm{r}_{ji}]$, $\bm{q'}=\textbf{W}_{6}[\bm{v}_i,\bm{q}]$ and $[\cdot,\cdot]$ denotes concatenation operation.

\paragraph{Intra-Modal Graph Convolution.}
Given the node and edge attention weights learned in Eq. {\ref{eq:node attention}} and Eq. \ref{edge attention}, the node representations of each layer of graphs are updated following the message-passing framework~\cite{gilmer2017neural}. We gather the neighborhood information and update the representation of $v_i$ as:
\begin{align}
&\bm{m}_i=\sum_{j\in\mathcal{N}_i}\beta_{ji}\bm{v'}_{j} \\
&\hat{\bm{v}}_i=\text{ReLU}(\textbf{W}_7[\bm{m}_i,\alpha_i\bm{v}_i])\label{eq:4}
\end{align}
where $\Ncal_i$ is the neighborhood set of node $v_i$.

We conduct the above intra-modal knowledge selection on $\Gcal^V$, $\Gcal^S$ and $\Gcal^F$ independently and obtain the updated node representations, denoted as $\{\hat{\bm{v}}_i^V\}_{i=1}^{\Ncal^V}$, $\{\hat{\bm{v}}_i^S\}_{i=1}^{\Ncal^S}$ and $\{\hat{\bm{v}}_i^F\}_{i=1}^{\Ncal^F}$ accordingly. 

\subsection{Cross-Modal Knowledge Reasoning}
\label{subsection:3.2}
To answer the question correctly, we fully consider the complementary evidence from visual, semantic and factual information. Since the answer comes from one entity in the fact graph, we gather complementary information from visual graph and semantic graph to fact graph by cross-modal convolutions, including \textit{visual-to-fact convolution} and \textit{semantic-to-fact convolution}. Finally, a \textit{fact-to-fact aggregation} is performed on the fact graph to reason over all the entities and form a global decision.

\paragraph{Visual-to-Fact Convolution.} 
For the entity $v_i^F$ in fact graph, the attention value of each node $v_j^V$ in the visual graph w.r.t. $v_i^F$ is calculated under the guidance of question:
\begin{equation}\label{eq:VtoF}
\gamma^{\textit{\text{V-F}}}_{ji}=\softmax(\bm{w}_c\tanh(\textbf{W}_8\hat{\bm{v}}^V_j+\textbf{W}_9[\hat{\bm{v}}^F_i,\bm{q}]))
\end{equation}

The complementary information $\bm{m}^\textit{\text{V-F}}_i$ from visual graph for $v_i^F$ is computed as:
\begin{equation}\label{eq:6}
\bm{m}^\textit{\text{V-F}}_i=\sum_{j\in\Ncal^V}\gamma^\textit{\text{V-F}}_{ji}\hat{\bm{v}}^V_j
\end{equation}

\paragraph{Semantic-to-Fact Convolution.}
The complementary information $\bm{m}^\textit{\text{S-F}}_i$ from the semantic graph is computed in the same way as in Eq. \ref{eq:VtoF} and Eq. \ref{eq:6}.

Then we fuse the complementary knowledge for $v_i^F$ from three layers of graphs via a gate operation:
\begin{align}
\label{eq:gate}
&\bm{gate}_i=\sigma(\textbf{W}_{10}[\bm{m}^\textit{\text{V-F}}_i, \bm{m}^\textit{\text{S-F}}_i, \hat{\bm{v}}_i^F])\\
&\widetilde{\bm{v}}^{F}_i=\textbf{W}_{11}(\bm{gate}_i\circ[\bm{m}^\textit{\text{V-F}}_i, \bm{m}^\textit{\text{S-F}}_i, \hat{\bm{v}}_i^F])
\end{align}
where $\sigma$ is sigmoid function and ``$\circ$'' denotes element-wise product. 

\paragraph{Fact-to-Fact Aggregation.} 
Given a set of candidate entities in the fact graph $\Gcal^F$, we aim to globally compare all the entities and select an optimal one as the answer. Now the representation of each entity in the fact graph gathers question-oriented information from three modalities. To jointly evaluate the possibility of each entity, we perform the attention-based graph convolutional network similar to Fact-to-Fact Convolution introduced in Section \ref{subsec:Intra} to aggregate information in the fact graph and obtain the transformed entity representations. 

We iteratively perform intra-modal knowledge selection and cross-modal knowledge reasoning in multiple steps to obtain the final entity representations. After $T$ steps, each entity representation $\widetilde{\bm{v}}^{F(T)}_i$ captures the structural information within $T$-hop neighborhood across three layers.

\subsection{Learning}
The concatenation of entity representation $\widetilde{\bm{v}}^{F(T)}_i$ and question embedding $\bm{q}$ is passed to a binary classifier  to predict its probability as the answer , \textit{i.e.} $\hat{y}_i=p_\theta([\widetilde{\bm{v}}^{F(T)}_i,\bm{q}])$. We apply the binary cross-entropy loss in the training process:
\begin{equation}
l_n=-\sum_{i\in \Ncal^F}\big[a\cdot{y}_i\ln \hat{y}_i + b\cdot(1-{y}_i)\ln(1-\hat{y}_i)\big]
\end{equation}
where ${y}_i$ is the ground truth label for $v_i^F$ and $a, b$ represent loss function weights for positive and negative samples respectively. The entity with the largest probability is selected as the final answer.

\section{Experiments}

\begin{table}
  \centering
	\setlength{\abovecaptionskip}{5pt}
	\setlength{\belowcaptionskip}{-1mm}
	\newcommand{\tabincell}[2]{\begin{tabular}{@{}#1@{}}#2\end{tabular}}
	\resizebox{\columnwidth}{!}{\scriptsize
		\begin{tabular}{l|cc}
			\hline
			\multirow{2}*{\bf Method} & \multicolumn{2}{c}{{\bf Overall Accuracy} }\\
			\cline{2-3}  & {\bf top-1} & {\bf top-3}\\
			\hline
			LSTM-Question+Image+Pre-VQA & $24.98$ & $40.40$ \\
			
			Hie-Question+Image+Pre-VQA& $43.14$ & $59.44$ \\
			
			FVQA (top-3-QQmaping) & $56.91$ & $64.65$ \\
			
			FVQA (Ensemble) & $58.76$ & - \\
			
			Straight to the Facts (STTF) & $62.20$ & $75.60$ \\
			
			Reading Comprehension & $62.96$ & $70.08$ \\
			
			Out of the Box (OB) &$69.35$&$80.25$\\
			\hline
			Human&$77.99$&-\\
			\hline
			{\bf Mucko}&$\bm {73.06}$&$\bm{ 85.94}$\\
			
			\hline
		\end{tabular}
	}
	\caption{State-of-the-art comparison on FVQA dataset.}
	\label{table:overall}
\end{table}

\begin{table}
  \centering
	\setlength{\abovecaptionskip}{5pt}
	\setlength{\belowcaptionskip}{-1mm}
	 \renewcommand{\arraystretch}{0.9}
	\resizebox{\columnwidth}{!}{\scriptsize
		\begin{tabular}{l|l|cc}
			\hline
			\multicolumn{2}{l|}{\multirow{2}*{\bf Method}}& \multicolumn{2}{c}{{\bf Overall Accuracy} }\\
			\cline{3-4}
			\multicolumn{2}{l|}{~}  & {\bf top-1} & {\bf top-3}\\
			\hline
			\multicolumn{2}{l|}{ {\bf Mucko} (full model)}& ${\bf73.06}$ & ${\bf85.94}$ \\
			\hline
			
			1& w/o Semantic Graph & $71.28$ & $82.76$ \\
			2&w/o Visual Graph& $69.12$ & $78.05$ \\
			3& w/o Semantic Graph \& Visual Graph & $20.43$ &$29.10$ \\
			\hline
			4& S-to-F Concat.& $67.82$ & $76.65$ \\
			5& V-to-F Concat. & $69.93$ & $80.12$ \\
			6& V-to-F Concat. \& S-to-F Concat. &$70.68$ & $82.04$ \\
			\hline
			7& w/o relationships &${ 72.10}$&${ 83.75}$\\
			\hline
		\end{tabular}
	}
	\caption{Ablation study of key components of Mucko.}
	\label{table:ablation}
\end{table}

\paragraph{Dataset.} We evaluate Mucko on the FVQA dataset \cite{wang2018fvqa}. It consists of 2,190 images, 5,286 questions and a knowledge base of 193,449 facts. Facts are constructed by extracting top visual concepts 
in the dataset and querying these concepts in 
WebChild, ConceptNet and DBPedia. \footnote{We provide more experimental results on OK-VQA and Visual7W+KB in supplementary materials.}
\paragraph{Evaluation Metrics.} We follow the metrics in \cite{wang2018fvqa} to evaluate the performance. The top-1 and top-3 accuracy is calculated for each method. The averaged accuracy of 5 test splits is reported as the overall accuracy. 
\paragraph{Implementation Details.} We select the top-10 dense captions according to their confidence. The max sentence length of dense captions and the questions is set to 20. The hidden state size of all the LSTM blocks is set to 512. 
We set $a=0.7$ and $b=0.3$ in the binary cross-entropy loss. 
Our model is trained by Adam optimizer with 20 epochs, where the mini-batch size is 64 and the dropout ratio is 0.5.
Warm up strategy is applied for 2 epochs with initial learning rate $1\times 10^{-3}$ and warm-up factor $0.2$. 
Then we use cosine annealing learning strategy with initial learning rate $\eta_{max}=1\times 10^{-3}$ and termination learning rate $\eta_{min}=3.6\times10^{-4}$ for the rest epochs. 

\begin{figure*}[t]
	\centering
	\setlength{\abovecaptionskip}{2pt}
	\setlength{\belowcaptionskip}{-4mm}
	\includegraphics[width=\textwidth]{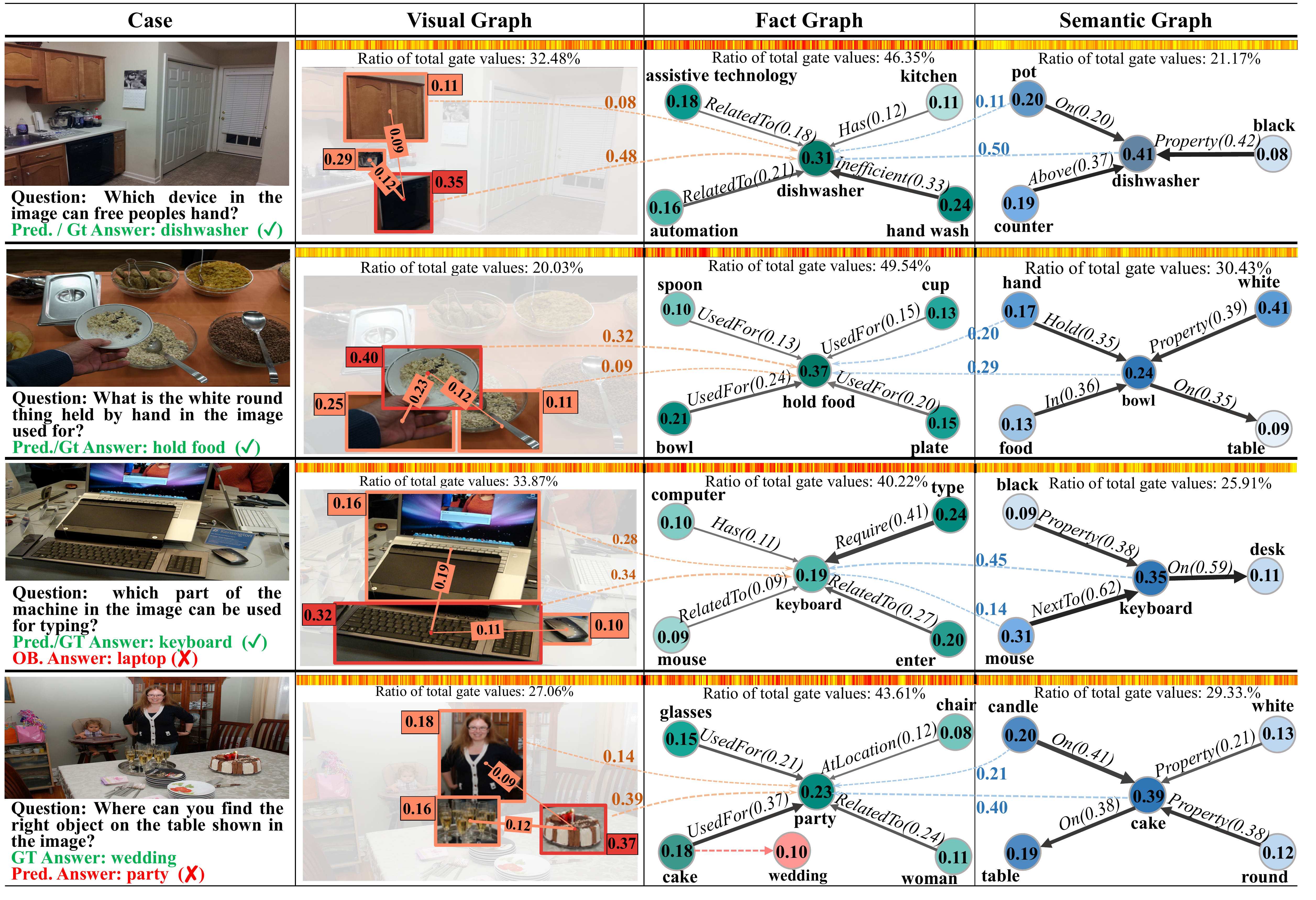}
	\caption{\small{Visualization for Mucko. Visual graph highlights the most relevant subject (red box) according to attention weights of each object ($\alpha^V$ in Eq. \ref{eq:node attention}) and the objects (orange boxes) with top-2 attended relationships ($\beta^V$ in Eq. \ref{edge attention}). Fact graph shows the predicted entity (center node) and its top-4 attended neighbors ($\alpha^F$ in Eq. \ref{eq:node attention}). Semantic graph shows the most relevant concept (center node) and its up to top-4 attended neighbors ($\alpha^S$ in Eq. \ref{eq:node attention}). Each edge is marked with attention value ($\beta^{F/S}$ in Eq. \ref{edge attention}). Dash lines represent visual-to-fact convolution (orange) and semantic-to-fact convolution weights (blue) of the predicted entity ($\gamma^{\textit{\text{V-F}}},\gamma^{\textit{\text{S-F}}}$ in Eq. \ref{eq:VtoF}).
	The thermogram on the top visualizes the gate values ($\bm{gate}_i$ in Eq. \ref{eq:gate}) of visual embedding (left), entity embedding (middle) and semantic embedding (right). 
	}} 
	\label{fig:case}
\end{figure*}

\subsection{Comparison with State-of-the-Art Methods}

Table \ref{table:overall} shows the comparison of Mucko with state-of-the-art models, including CNN-RNN based approaches \cite{wang2018fvqa}, \textit{i.e.} LSTM-Question+Image+Pre-VQA and Hie-Question+Image+Pre-VQA, semantic parsing based approaches \cite{wang2018fvqa}, \textit{i.e.} FVQA (top-3-QQmaping) and FVQA (Ensemble), learning-based approaches, \textit{i.e.} Straight to the Facts (STTF) \cite{narasimhan2018straight} and Out of the Box (OB) \cite{narasimhan2018out}, and Reading Comprehension based approach \cite{li2019visual}. Our model consistently outperforms all the approaches on all the metrics and achieves 3.71\% boost on top-1 accuracy and 5.69\% boost on top-3 accuracy compared with the state-of-the-art model. The model OB is most relevant to Mucko in that it leverages graph convolutional networks to jointly assess all the entities in the fact graph. However, it introduces the global image features equally to all the entities without selection. By collecting question-oriented visual and semantic information via modality-aware heterogeneous graph convolutional networks, our model gains remarkable improvement.

\subsection{Ablation Study}

In Table \ref{table:ablation}, we shows ablation results to verify the contribution of each component in our model.
 (1) In models `1-3', we evaluate the \textbf{influence of each layer of graphs} on the performance. 
We observe that the top-1 accuracy of `1' and `2' respectively decreases by 1.1\% and 3.94\% compared with the full model, which indicates that both semantic and visual graphs are beneficial to 
provide valuable evidence for answer inference. Thereinto, the visual information has greater impact than the semantic part. When removing both semantic and visual graphs, `3' results in a significant decrease.
(2) In models `4-6', we assess the \textbf{effectiveness of the proposed cross-modal graph convolutions}. `4', `5' and `6' respectively replace the `Semantic-to-Fact Conv.' in `2', `Visual-to-Fact Conv.' in `1' and both in full model by concatenation, \textit{i.e.} concatenating the mean pooling of all the semantic/visual node features with each entity feature.
The performance decreases when replacing the convolution from either S-to-F or V-to-F, or both simultaneously, which proves the benefits of cross-modal convolution in gathering complementary evidence from different modalities. (3) We evaluate the \textbf{influence of the relationships} in the heterogeneous graph. We omit the relational features $\bm{r}_{ij}$ in all the three layers in `7' and the performance decreases by nearly 1\% on top-1 accuracy. It proves the benefits of relational information, though it is less influential than the modality information. 

\subsection{Interpretability}
Our model is interpretable by visualizing the attention weights and gate values in the reasoning process.
From case study in Figure \ref{fig:case}, we conclude with the following three insights: 
\textbf{(1) Mucko is capable to reveal the knowledge selection mode.} The first two examples indicate that Mucko captures the most relevant visual, semantic and factual evidence as well as complementary information across three modalities. In most cases, factual knowledge provides predominant clues compared with other modalities according to gate values because FVQA relies on external knowledge to a great extent. Furthermore, more evidence comes from the semantic modality when the question involves complex relationships. 
For instance, 
the second question involving the relationship between `hand' and `while round thing' needs more semantic clues.
\textbf{(2) Mucko has advantages over the state-of-the-art model.} The third example compares the predicted answer of OB with Mucko. Mucko collects relevant visual and semantic evidence to make each entity discriminative enough for predicting the correct answer while OB failing to distinguish representations of `laptop' and `keyboard' without feature selection.
\textbf{(3) Mucko fails when multiple answers are reasonable for the same question.} Since both `wedding' and `party' may have cakes, the predicted answer `party'  in the last example is  reasonable from human judgement.

\begin{table}
	\centering
	\renewcommand{\arraystretch}{1}
	\setlength{\abovecaptionskip}{5pt}
	\setlength{\belowcaptionskip}{-2mm}
	\resizebox{0.9\columnwidth}{!}{\scriptsize
		\begin{tabular}{l|rrrr}
			\hline
			{\bf \#Retrieved facts}& {\bf @50} &{\bf @100} &{\bf @150} &{\bf @200} \\
			
			\hline
			\multirow{2}*{\shortstack{\bf Rel@1 (top-1 accuracy) \\\bf Rel@1 (top-3 accuracy)}} & 55.56 & 70.62 & 65.94 & 59.77 \\
			~ & 64.09 & 81.95 & 73.41 & 66.32 \\
			\hline
			\multirow{2}*{\shortstack{\bf Rel@3 (top-1 accuracy) \\\bf Rel@3 (top-3 accuracy)}}  & 58.93 & {\bf73.06}&70.12&65.93 \\
			~& 68.50& {\bf85.94} & 81.43 & 74.87 \\
			\hline
		\end{tabular}
	}
	\caption{Overall accuracy with different number of retrieved candidate facts and different number of relation types.}
	\label{table:factrecall}
\end{table}

\begin{table}
	\centering
	\setlength{\abovecaptionskip}{5pt}
	\setlength{\belowcaptionskip}{-4mm}
	\resizebox{0.7\columnwidth}{!}{\scriptsize
	\begin{tabular}{l|ccc}
		\hline
		{\bf \#Steps }  & 1 & 2 & 3\\
		\hline
		{\bf top-1 accuracy}&62.05 & {\bf 73.06} & 70.43 \\
		{\bf top-3 accuracy} & 71.87 & {\bf85.94} & 81.32 \\
		\hline
	\end{tabular}
		}
	\caption{Overall accuracy with different number of reasoning steps.}
	\label{table:step}

\end{table}

\subsection{Parameter Analysis}

In Table \ref{table:factrecall}, we vary the number of retrieved candidate facts and relation types for candidate filtering. We achieve the highest downstream accuracy with top-100 candidate facts and top-3 relation types.
In Table 4, we evaluate the influence of different number of reasoning steps $T$. 
We find that two reasoning steps achieve the best performance. We use the above settings in our full model.

\section{Conclusion}
In this paper, we propose Mucko for visual question answering requiring external knowledge, which focuses on multi-layer cross-modal knowledge reasoning.
We novelly depict an image by a heterogeneous graph with multiple layers of information corresponding to visual, semantic and factual modalities. We propose a modality-aware heterogeneous graph convolutional network to select and gather intra-modal and cross-modal evidence iteratively.
Our model outperforms the state-of-the-art approaches remarkably and obtains interpretable results on the benchmark dataset.

\section*{Acknowledgements}
This work is supported by the National Key Research and Development Program (Grant No.2017YFB0803301).

\clearpage
\section{Supplementary Materials}
We also conduct extensive experiments on another two large-scale knowledge-based VQA datasets: OK-VQA \cite{marino2019ok} and Visual7W+KB \cite{li2017incorporating} to evaluate performance of our proposed model. In this section, we first briefly review the dataset and then report the performance of our proposed method comparing with several baseline models.

\subsection{Datasets}

\paragraph{Visual7W+KB:} The Visual7W dataset \cite{zhu2016visual7w} is built based on a subset of images from Visual Genome \cite{krishna2017visual}, which includes questions in terms of (what, where, when, who, why, which and how) along with the corresponding answers in a multi-choice format. However, most of questions of Visual7W solely base on the image content which don't require external knowledge. Furthermore, \cite{li2017incorporating} generated a collection of knowledge-based questions based on the test images in Visual7W by filling a set of question-answer templates that need to reason on both visual content and external knowledge. We denoted this dataset as Visual7W+KB in our paper. In general, Visual7W+KB consists of 16,850 open-domain question-answer pairs based on 8,425 images in Visual7W test split. Different from FVQA, Visual7W+KB uses ConceptNet to guide the question generation but doesn't provide a task-specific knowledge base. In our work, we also leverage ConceptNet to retrieve the supporting knowledge and select one entity as the predicted answer. 

\paragraph{OK-VQA:} \cite{marino2019ok} proposed the Outside Knowledge VQA (OK-VQA) dataset, which is the largest knowledge-based VQA dataset at present. Different from existing datasets, the questions in OK-VQA are manually generated by MTurk workers, which are not derived from specific knowledge bases. Therefore, it requires the model to retrieve supporting knowledge from open-domain resources, which is much closer to the general VQA but more challenging for existing models. OK-VQA contains 14,031 images which are randomly collected from MSCOCO dataset \cite{lin2014microsoft}, using the original 80k-40k training and validation splits as train and test splits. OK-VQA contains 14,055 questions covering a variety of knowledge categories such as science \& technology, history, and sports.

\subsection{Experimental results on Visual7W+KB}
The comparison of state-of-the-art models on Visual7W-KB dataset is shown in the Table \ref{table:V7W_SOTA}. The compared baselines contains two sets, i.e. memory-based approaches and a graph-based approach. The memory-based approaches \cite{li2017incorporating} include KDMN-NoKnowledge (w/o external knowledge), KDMN-NoMemory (attention-based knowledge incorporation), KDMN (dynamic memory network based knowledge incorporation) and KDMN-Ensemble (several KDMN models based ensemble model). We also test the performance of Out of the Box (OB) \cite{narasimhan2018out} on Visual7W-KB and report the results in Table \ref{table:V7W_SOTA}.

As consistent with the results on FVQA, we achieve a significant improvement (7.98\% on top-1 accuracy and 13.52\% on top-3 accuracy ) over state-of-the-art models. Note that our proposed method is an single-model, which outperforms the existing ensembled model \cite{li2017incorporating}. 
\begin{table}[tp]
	\centering
	\newcommand{\tabincell}[2]{\begin{tabular}{@{}#1@{}}#2\end{tabular}}
	\resizebox{\columnwidth}{!}{\scriptsize
		\begin{tabular}{l|cc}
			\hline
			\multirow{2}*{\bf Method} & \multicolumn{2}{c}{{\bf Overall Accuracy} }\\
			\cline{2-3}  & {\bf top-1} & {\bf top-3}\\
			\hline
			KDMN-NoKnowledge \cite{li2017incorporating}  & $45.1$ & - \\
			
			KDMN-NoMemory \cite{li2017incorporating}&$51.9$&-\\
			
			KDMN \cite{li2017incorporating}&$57.9$&-\\
			
			KDMN-Ensemble \cite{li2017incorporating}&$60.9$&-\\
	        Out of the Box (OB) \cite{narasimhan2018out}&$57.32$&$71.61$\\
			\hline
    
			{\bf Mucko (ours)}&$\bm {68.88}$&$\bm{ 85.13}$\\
			
			\hline
		\end{tabular}
	}
	\caption{State-of-the-art comparison on Visual7W+KB dataset.}
	\label{table:V7W_SOTA}
\end{table}

\subsection{Experimental results on OK-VQA}
We also report the performance on the challenging OK-VQA dataset in Table \ref{table:OK-VQA_SOTA}. We compare our model with three kinds of existing models, including current state-of-the-art VQA models, knowledge-based VQA models and ensemble models. The VQA models contain Q-Only \cite{marino2019ok}, MLP \cite{marino2019ok}, BAN \cite{kim2018bilinear}, MUTAN\cite{kim2018bilinear}. The knowledge-based VQA models \cite{marino2019ok} consist of ArticleNet (AN), BAN+AN and MUTAN+AN. The ensemble models \cite{marino2019ok}, i.e. BAN/AN oracle and MUTAN/AN oracle,  simply take the raw ArticleNet and VQA model predictions, taking the best answer (comparing to ground truth) from either. 

Our model consistently outperforms all the compared models on the overall performance. Even the state-of-the-art models (BAN and MUTAN) specifically designed for VQA tasks, they get inferior results compared with ours. This indicates that general VQA task like OK-VQA cannot be simply solved by a well-designed model, but requires the ability to incorporate external knowledge in an effective way. Moreover, our model outperforms knowledge-based VQA models including both single models (BAN+AN and MUTAN+AN) and ensemble models (BAN/AN oracle and MUTAN/AN oracle), which further proves the advantages of our proposed multi-layer heterogeneous graph representation and cross-modal heterogeneous graph reasoning.

\begin{table}[tp]
\centering
\resizebox{\columnwidth}{!}{\scriptsize
\begin{tabular}{l|cc}
\hline
\multirow{2}*{\bf Method} & \multicolumn{2}{c}{{\bf Overall Accuracy}}\\
\cline{2-3}   & {\bf top-1} & {\bf top-3}\\
\hline
Q-Only \cite{marino2019ok} & 14.93 &-\\
MLP \cite{marino2019ok} & 20.67 &- \\
BAN \cite{kim2018bilinear} & 25.17&-  \\
MUTAN \cite{ben2017mutan} & 26.41&-  \\
ArticleNet (AN) \cite{marino2019ok} & 5.28&- \\
BAN + AN \cite{marino2019ok}& 25.61 & - \\
MUTAN + AN \cite{marino2019ok}& 27.84 & -  \\
BAN/AN oracle  \cite{marino2019ok}& 27.59  & - \\
MUTAN/AN oracle \cite{marino2019ok}& 28.47  & - \\ 
\hline
\textbf{Mucko (ours)} & $\bm{29.20}$&$\bm{30.66} $ \\\hline
\end{tabular}
}
\caption{State-of-the-art comparison on OK-VQA dataset.}
\label{table:OK-VQA_SOTA}
\end{table}

\clearpage
\bibliographystyle{named}
\bibliography{ijcai20}

\end{document}